\documentclass[10pt, conference]{IEEEtran}
\IEEEoverridecommandlockouts
\IEEEpubid{\makebox[\columnwidth]{978-1-6654-3274-0/21/\$31.00~\copyright2021 IEEE \hfill} \hspace{\columnsep}\makebox[\columnwidth]{ }}
\usepackage{cite}
\usepackage{amsmath,amssymb,amsfonts}
\usepackage{graphicx}
\usepackage{textcomp}
\usepackage{xcolor}
\usepackage{bbm}
\usepackage{stfloats}
\usepackage{times}
\usepackage{epsfig}
\usepackage{graphicx}
\usepackage{amsmath}
\usepackage{amssymb}
\usepackage[flushleft]{threeparttable} 
\usepackage{booktabs} 
\usepackage{multirow}
\usepackage{algorithm}
\usepackage{algpseudocode}

\DeclareMathOperator*{\argmax}{arg\,max}
\DeclareMathOperator*{\argmin}{arg\,min}

\def\BibTeX{{\rm B\kern-.05em{\sc i\kern-.025em b}\kern-.08em
    T\kern-.1667em\lower.7ex\hbox{E}\kern-.125emX}}
    
\begin{document}

\title{A3C-S: Automated Agent Accelerator Co-Search towards Efficient Deep Reinforcement Learning
}

\author{\IEEEauthorblockN{Yonggan Fu}
\IEEEauthorblockA{\textit{Rice University} \\
yf22@rice.edu}
\and
\IEEEauthorblockN{Yongan Zhang}
\IEEEauthorblockA{\textit{Rice University} \\
yz87@rice.edu}
\and
\IEEEauthorblockN{Chaojian Li}
\IEEEauthorblockA{\textit{Rice University} \\
cl114@rice.edu}
\and
\IEEEauthorblockN{Zhongzhi Yu}
\IEEEauthorblockA{\textit{Rice University} \\
zy42@rice.edu}
\and
\IEEEauthorblockN{Yingyan Lin}
\IEEEauthorblockA{\textit{Rice University} \\
yingyan.lin@rice.edu}
}

\maketitle

\begin{abstract}

Driven by the explosive interest in applying deep reinforcement learning (DRL) agents to numerous real-time control and decision-making applications, there has been a growing demand to deploy DRL agents to empower daily-life intelligent devices, while the prohibitive complexity of DRL stands at odds with limited on-device resources. In this work, we propose an Automated Agent Accelerator Co-Search (A3C-S) framework, which to our best knowledge is the first to automatically co-search the optimally matched DRL agents and accelerators that maximize both test scores and hardware efficiency. Extensive experiments consistently validate the superiority of our A3C-S over state-of-the-art techniques.


\end{abstract}

\section{Introduction}
\label{sec:intro}

Recent successes in deep reinforcement learning (DRL)~\cite{mnih2015human}, which integrates reinforcement learning (RL) and deep neural networks (DNNs), have triggered tremendous enthusiasm in developing and deploying DRL-powered intelligence into numerous inference and control applications, including robotics and autonomous vehicles.
Many of them, such as autonomous vehicles, require real-time control and decision-making policies for which the DRL agents have to derive real-time policies using real-time data for dynamic systems.
However, real-time control and decision-making for DRL can be prohibitively challenging in many real-world applications due to DRL's integrated complex DNNs and edge devices' constrained resources, calling for DRL-powered intelligent solutions that favor both test scores and hardware efficiency.    

To address the large gap between the growing need for on-device DRL and DRL's prohibitive complexity, emerging network and accelerator co-exploration (NACoS) methods~\cite{jiang2019hardware,li2020edd, zhang2020dna} are promising as they can boost DNN acceleration efficiency. However, directly applying NACoS methods to design DRL agents can easily fail due to the commonly observed vulnerability and instability of DRL training, which occur with a high variance ~\cite{cheng2019control}. Furthermore, such instability will be further exacerbated when considering differentiable NAS (DNAS) based NACoS methods, which requires competitively low search cost and thus can enable efficient navigation over the large DRL agent and accelerator joint space, because the success of DNAS requires unbiased gradient estimation with low variance. To this end, we aim to develop a novel NACoS framework dedicated to DRL agent acceleration to promote fast development and highly efficient DRL-powered solutions. Specifically, we make the following contributions:

\begin{itemize}
    \item We propose an Automated Agent Accelerator Co-Search (A3C-S) framework, which to our best knowledge is the first to automatically co-search the optimally matched DRL agents and accelerators that maximize both test scores and hardware efficiency.
    
    \item A3C-S integrates and demonstrates the first DNAS search dedicated to DRL that features a novel distillation mechanism to effectively stabilize agent search, despite the instability of DRL training.
    
    \item A3C-S incorporates 
    a parameterized micro-architecture with over $10^{27}$ searchable choices of accelerators and dataflows to enable a differentiable search for DRL agent accelerators. A3C-S is generic and can be applied on top of different accelerator templates. 
        
     \item Through FPGA measurements, extensive experiments and ablation studies validate A3C-S's effectiveness in generating efficient DRL agents/accelerators that consistently outperform state-of-the-art (SOTA) agents/accelerators. 

\end{itemize}

\section{Related works}
\label{sec:related_work}

\textbf{Deep reinforcement learning.}
DRL integrates traditional RL algorithms with DNNs 
to handle higher-dimensional and more complex problems, e.g., 
 DQN~\cite{mnih2015human} introduces DNNs to Q-Learning and \cite{mnih2016asynchronous} utilizes DNNs to model both the actor and critic in AC-based DRL. 
More DRL works can be found in ~\cite{arulkumaran2017deep}. Despite DRL's promising success, automating the design of DRL agents has not yet been explored, while there is a growing need for fast development of DRL-powered solutions.

\textbf{Hardware-aware neural architecture search.}
NAS has been widely adopted to automate the design of efficient DNNs. To tackle the prohibitive search cost of previous RL-based NAS methods, DNAS~\cite{fu2020autogandistiller,wu2019fbnet} has gained more attention thanks to its excellent search efficiency.
However, the possibility of applying NAS or DNAS to DRL hasn't been explored.

\textbf{DNN accelerator design.}
SOTA DNN accelerators~\cite{smartexchange,9138916} tackle the prohibitive complexity of DNNs through novel micro-architectures/dataflows to maximize data reuses, and thus improve acceleration efficiency. Early works require experts' manual design, and thus were time-consuming.
Later, various design flow and automation tools~\cite{venkatesanmagnet,xu2020autodnnchip} were developed. However, they mostly explored DNN acceleration, leaving automated DRL accelerator design unexplored.

\textbf{Network and accelerator co-search.}
Jointly exploring DNNs and their accelerators is very promising towards efficient DNN solutions, as shown in pioneering works ~\cite{jiang2019hardware,li2020edd}, where the former suffers from large search time and the latter considers a limited search space. 
These works motivate us to explore the joint search for DRL agents and their accelerators to maximize the test scores and hardware efficiency.

\section{Preliminaries of DRL}
\label{sec:preliminary}

Here we describe the preliminaries of DRL.
RL can be viewed as a Markov Decision Process (MDP) determined by a tuple $(S, A, T, R, \gamma)$, where $S$ is the state space, $A$ is the action space, $T(s'|s,a)$ is the transition probability of ending up in state $s'$ when executing the action $a$ in the state $s$, $R$ is the reward function, and $\gamma$ is a discount factor. A policy $ \pi({s_{t}, a_{t}}) = p(a_{t} | s_{t} ) $ defines the probability that an agent in the MDP executes the action $a$ in the state $s$. In particular, the agent performs the action $ a_{t} \in A $ sampled from the policy $ \pi({s_{t}, a_{t}})$ at each time step $ t $ and the state $ s_{t} \in S $, leading to the next state $ s_{t+1} \in S $ and acquiring a reward $ r_{t} $. 




In DRL, a policy is parameterized by $ \theta_{\pi}$, i.e., the weights of a DNN, and the agent's goal is learning an optimal policy to maximize the expected cumulative reward:
\vspace{-0em}
\begin{equation}
\label{eq:rl_policy_objective}
\theta_{\pi}^{*} = \underset{\theta_{\pi}}{\argmax} \,\, J(\pi(\cdot | \theta_{\pi})) = \underset{\theta_{\pi}}{\argmax} \,\, \mathbb{E}_{\pi} \left[ \sum_{t=0}^{H} \gamma^t r_{t} \right]
\end{equation}

\noindent where $ H $ is the time horizon, and $ \gamma $ is a 
discount factor.

To solve the optimization problem in Eq.~\ref{eq:rl_policy_objective}, stochastic policy gradient methods~\cite{sutton2000policy} are widely adopted where $\nabla_{\theta_{\pi}} J(\pi(\cdot | \theta_{\pi})) $:
\begin{equation}
\label{eq:rl_gradient}
\nabla_{\theta_{\pi}} J(\pi(\cdot | \theta_{\pi})) = \mathbb{E}_{\pi} \left[ \sum_{t=0}^{H} \delta_{t} \nabla_{\theta_{\pi}} \log(\pi(a_t, s_t | \theta_{\pi})) \right]
\end{equation}

\noindent In this work, we adopt the temporal difference error (td-error) $\delta_{t} = r_{t} + \gamma V_{\pi}(s_{t+1}) - V_{\pi}(s_{t})$ to reduce the variance in policy gradients, where $ V_{\pi}(s) = \mathbb{E}_{\pi} \left[ \sum_{t=0}^{H} \gamma^{t} r_{t} | s_0 = s \right] $ is the value function which estimates the expected cumulative reward of the policy $\pi$ starting from the state $s$. Since the value function $V_{\pi}$ is hard to estimate, AC-based DRL methods~\cite{konda2000actor} parameterize the value function (i.e., the critic) with the learnable parameter $ \theta_{v} $, i.e. the weights of a DNN in DRL. The objective of $ \theta_{v} $ is to minimize the td-error of the estimated value between consecutive states:
\vspace{-0.5em}
\begin{equation}
\label{eq:rl_value_objective}
\theta_{v}^{*}  = \underset{\theta_{v}}{\argmin} \,\, \mathbb{E}_{\pi} \left[ \sum_{t=0}^{H} \frac{1}{2} \left( r_{t} + \gamma V_{\pi}(s_{t+1} | \theta_{v}) - V_{\pi}(s | \theta_{v}) \right)^2 \right] \\ 
\end{equation}

\noindent Therefore, in the AC framework, the actor and critic parameterized by $\theta_{\pi}$ and $ \theta_{v} $, respectively can be iteratively updated to lead the agent towards an optimal policy.

\section{The Proposed A3C-S Framework}
There exist three main challenges in designing A3C-S: (1) the huge joint search space, (2) the non-differentiable accelerator parameters and the gap between DNAS's required layer-wise hardware-cost penalty and the optimal accelerators' dependency on all the layers, and (3) the training instability with a high variance of DRL which may lead to the failure of applying NAS to search DRL agents. 
In this section, we first introduce the component techniques of A3C-S in Sec.~\ref{sec:co-search} that tackles the first two challenges with a novel differentiable search strategy and a differentiable accelerator search engine, and then A3C-S's AC-distillation mechanism  in Sec.~\ref{sec:ac-distillation} to tackle the third challenge.

\subsection{A3C-S: the co-search pipeline}
\label{sec:co-search}

\textbf{A3C-S formulation.} We formulate A3C-S as below:

\vspace{-1em}
\begin{align} 
    \begin{split}
     \min \limits_{\theta_{\pi}, \theta_{v}, \alpha} \,\, & L_{task}( \theta_{\pi}, \theta_{v}, net(\alpha))+\lambda L_{cost}(hw(\phi^*), net(\alpha)) \label{eq:update_alpha} 
    \end{split} \\
    \begin{split}
    & s.t. \quad \phi^* = \underset{\phi}{\arg\min} \,\, L_{cost}(hw(\phi), net(\alpha)) \label{eq:update_hw}
    \end{split}
\end{align}
\vspace{-1.em}

\noindent where $\alpha$ and $\phi$ are the variables maintaining the probability of choosing different (1) network operators and (2) accelerator parameters, with $\theta_{\pi}$ and $\theta_{v}$ being the supernet weights of the actor and critic in a AC-based DRL, respectively, $L_{task}$ and $L_{cost}$ are the task loss (see Sec.~\ref{sec:ac-distillation} for details) and the hardware-cost loss, respectively, and $net(\alpha)$ and $hw(\phi)$ denote the network and accelerator parameterized by $\alpha$ and $\phi$, respectively.   

\textbf{A3C-S's co-search pipeline.} During co-search, 
A3C-S starts by updating the accelerator parameters $\phi$ given the current network structure $net(\alpha)$, and then updates $\theta_{\pi}$, $\theta_{v}$, and $\alpha$ in the same iteration based on the accelerator $hw(\phi^*)$ resulting from the previous step. Our A3C-S adopts one-level optimization~\cite{xie2018snas} instead of bi-level optimization~\cite{liu2018darts}, considering that 
the one-step SGD approximation of bi-level optimization will lead to biased gradient estimation~\cite{he2020milenas} which can largely suffer from the high variance of DRL training evaluated in Sec.~\ref{sec:exp_nas}. The updates of $\alpha$ and $\phi$ follow: 

\vspace{-1em}
\begin{align}
    \begin{split}
    Forward: A^{l+1} = \sum_{i=1}^{N} GS_{hard}(\alpha_i^l) \, O_i(A^l) = O_{fw}^l(A^l)
    \label{eq:forward} 
    \end{split} \\
    \begin{split}
    Backward: \frac{\partial L_{task}}{\partial \alpha_i^l} = \sum^{K}_{k=1} \frac{\partial L_{task}}{\partial GS(\alpha_k^l)} \frac{\partial GS(\alpha_k^l)}{\partial \alpha_i^l} \\
    = \frac{\partial L_{task}}{\partial A^{l+1}} \sum^{K}_{k=1} O_k^l(A^l) \frac{\partial GS(\alpha_k^l)}{\partial \alpha_i^l}
    \label{eq:backward_task} 
    \end{split}
    \\
    \begin{split}
          \frac{\partial L_{cost}}{\partial \alpha_i^l} = \boldsymbol{\mathbbm{1}}(GS_{hard}(\alpha_i^l)=1) \, L^{\alpha_i^l}_{cost}(hw(\phi^*), net(\alpha_i^l))
    \label{eq:backward_cost} 
    \end{split}
\end{align}

\noindent where $A^{l}$ and $A^{l+1}$ are the feature maps of the $l$-th and $(l+1)$-th layer, respectively, $GS_{hard}$ is the hard Gumbel Softmax operator generating a one-hot output, i.e., only one operator $O_{fw}^l$ will be activated during forward, $N$ is the total number of operator choices, and $O_i^l$ is the $i$-th operator in the $l$-th layer parameterized by $\alpha_i^l$. Meanwhile, $GS$ is a Gumbel Softmax function and $K$ is the number of activated paths with the top $K$ probability, and similar to~\cite{cai2018proxylessnas}, $K\in(1,N)$ in A3C-S to control the computational cost. In Eq.~\ref{eq:backward_cost}, $\boldsymbol{\mathbbm{1}}$ is an indicator denoting whether $\alpha_i^l$ (i.e., the $i$-th operator in the $l$-th layer) is activated during forward.

\textbf{A3C-S's co-search strategy.} A3C-S integrates a novel search strategy to solve Eq.~\ref{eq:update_alpha} for effective yet efficient search to avoid memory explosion due to the large joint search spaces:

\underline{Single-path forward:} (see Eq.~\ref{eq:forward}) A3C-S adopts hard Gumbel Softmax sampling~\cite{jang2016categorical}, i.e., only the choice with the highest probability will be activated to narrow the gap between the supernet and the finally derived network thanks to the single-path property of hard Gumbel Softmax sampling.

\underline{Multi-path backward:} (see Eq.~\ref{eq:backward_task}) A3C-S activates multiple paths to approximate the gradients of $\alpha$ via Gumbel Softmax relaxation to balance the search efficiency (prefer fewer activated paths) and stability (prefer more activated paths), inspired by~\cite{cai2018proxylessnas} which targets DNAS for DNNs.

\underline{Hardware-cost penalty:} The network search in Eq.~\ref{eq:update_alpha} requires  layer-wise hardware-cost penalties assuming the layer-wise operators running on the final optimal accelerator $hw(\phi^*)$, which is not yet available at each co-search epoch as the optimal network is still unknown, i.e., the chicken-and-egg problem. To handle this, we approximate the layer-wise hardware-cost by assuming that the single-path network derived during each forward is close to the final derived network, since the network
operators that have higher probabilities are also more likely
to appear in the final optimal network.



\textbf{A3C-S's Differentiable accelerator search (DAS).}
EDD~\cite{li2020edd} made a pioneering effort to differentiably co-search the network and accelerator, yet their accelerator search space is limited to the parallel factor of their template, which can be analytically fused into their computational cost, whereas this is not always applicable to other naturally non-differentiable accelerator design knobs such as PE numbers and buffer allocation strategies. A more general search engine is desirable.

\underline{A3C-S's accelerator search algorithm:} We propose a general DAS engine to efficiently search for the optimal accelerator, including the micro-architectures and dataflows, given a DNN based on the single-path sampling in Eq.~\ref{eq:forward}, i.e.:


\vspace{-1.5em}
\begin{equation} \label{eq:hw_diff}
    \begin{split}
    \phi^*& = \,\, \underset{\phi}{\arg\min} \,\, \sum_{m=1}^{M} GS_{hard}(\phi^m) \, \hat{L}\\
    where \,\,\,\, \hat{L}&=L_{cost}(hw(\{GS_{hard}(\phi^m)\}), net(\alpha))
    \end{split}
\end{equation}

\noindent where $M$ is the number of accelerator parameters. Given the network $net(\alpha)$ which is the most likely network sampled during the single-path forward, the search engine utilizes hard Gumbel Softmax $GS_{hard}$ sampling on each design parameter $\phi^m$ to build an accelerator $hw(\{GS_{hard}(\phi^m)\})$ and penalize each sampled accelerator parameter with the overall hardware-cost $L_{cost}$ through relaxation in a differentiable manner.

\underline{A3C-S's accelerator template:} We adopt a parameterized accelerator template built upon a SOTA chunk-based pipeline micro-architecture~\cite{shen2017maximizing}. The accelerator template comprises multiple sub-accelerators (i.e., chunks) and executes DNNs in a pipeline fashion. In particular,
each chunk is assigned with multiple but not necessarily consecutive layers which are executed sequentially within the chunk. Similar to Eyeriss, each chunk consists of levels of buffers/memories (e.g., on-chip buffer and local register files) and processing elements (PEs) to facilitate data reuses and parallelism with searchable accelerator parameters, including PE interconnections (i.e., Network-on-chip), buffer sizes, and MAC operations’ scheduling and tiling (i.e., dataflows) (see more details in Sec.~\ref{sec:exp_setup}).

\subsection{A3C-S: the AC-distillation mechanism}
\label{sec:ac-distillation}

\textbf{Motivation.} Policy distillation~\cite{rusu2015policy} shows that the distillation from a teacher agent can effectively reduce the variance of gradient estimates and stabilize the training process of the student agent, motivating us to introduce a distillation mechanism to stabilize the DNAS process for DRL. However, vanilla policy distillation merely distills the policy without considering the value function which can play a critical role in both assisting the policy updates and reducing the variance of vanilla policy gradients. We conjecture that further distilling the value function from the teacher agent can better improve the training stability and the convergence. 

\textbf{A3C-S's AC-distillation.} In A3C-S, we propose an AC-distillation mechanism to distill  knowledge from both the actor and critic of a pretrained teacher agent to the student agent, where the two distillation losses for the actor and critic are:
\begin{equation}
\label{eq:loss_actor}
L_{actor}^{distill} = \mathbb{E}_{\pi} \left[ \sum_{t=0}^{H} \pi(a_t, s_t | \theta_{\pi}^{tea}) \,\,
log \frac{\pi(a_t, s_t | \theta_{\pi}^{tea})}{ \pi(a_t, s_t | \theta_{\pi}^{stu})} \right]
\end{equation}

\begin{equation}
\label{eq:loss_critic}
L_{critic}^{distill}  =  \mathbb{E}_{\pi} \left[ \sum_{t=0}^{H} \frac{1}{2} \left( V_{\pi}(s_{t} | \theta_{v}^{stu}) - V_{\pi}(s_{t} | \theta_{v}^{tea}) \right)^2  \right]
\end{equation}
\vspace{-0.2em}

\noindent where $\pi(a_t, s_t | \theta_{\pi}^{tea})$ and $\pi(a_t, s_t | \theta_{\pi}^{stu}) $ are the teacher and student actor, respectively, and $V_{\pi}(s_{t} | \theta_{v}^{tea})$ and $V_{\pi}(s_{t} | \theta_{v}^{stu})$ are the teacher and student critic, respectively. We adopt KL divergence to distill the knowledge from the teacher actor following~\cite{rusu2015policy} and the MSE loss as a soft constraint to enforce the student critic to mimic the estimated value of the teacher critic. The final objective during both search and training is:

\vspace{-1em}
\begin{equation}
\begin{split}
\label{eq:loss_total}
L_{task}  = \,\, & L_{policy} + L_{value} + \beta_{1}L_{entropy} \\ 
 + \,\, & \beta_{2} L_{actor}^{distill} + \beta_{3}L_{critic}^{distill}
\end{split}
\end{equation}

\noindent where $\beta_1$, $\beta_2$, and $\beta_3$ are the weighted coefficients.
Here $L_{policy}$ is the policy gradient loss as in~\cite{sutton2000policy}, $L_{value}$ is the value loss based on the td-error, and $L_{entropy}$ is the entropy loss on top of the policy to encourage exploration, i.e.:

\begin{equation}
\label{eq:loss_policy}
L_{policy}  = \mathbb{E}_{\pi} \left[ - \sum_{t=0}^{H} \delta_{t} \log(\pi(a_t, s_t | \theta_{\pi}^{stu})) \right]
\end{equation}
\vspace{-1em}
\begin{equation}
\label{eq:loss_value}
L_{value} = \mathbb{E}_{\pi} \left[ \sum_{t=0}^{H} \frac{1}{2} \left( r_{t} + \gamma V_{\pi}(s_{t+1} | \theta_{v}^{stu}) - V_{\pi}(s_t | \theta_{v}^{stu}) \right)^2 \right] \\
\end{equation}
\vspace{-1em}
\begin{equation}
\label{eq:loss_entropy}
L_{entropy} = \mathbb{E}_{\pi} \left[ \sum_{t=0}^{H} \pi(a_t, s_t | \theta_{\pi}^{stu}) \,\, log(\pi(a_t, s_t | \theta_{\pi}^{stu})) \right]
\end{equation}

The search algorithm of A3C-S is summarized in Alg.~\ref{alg:a3c-s}.

\begin{algorithm}[t!]
\caption{Automated Agent Accelerator Co-Search (A3C-S)}
\begin{algorithmic}
\small
\State Initialize the step counter $t\gets 1$
\Repeat
\State $t_{start} = t$
\State Get state $s_t$
\Repeat
\State Perform $a_t \sim \pi(a_t, s_t | \theta_{\pi}^{stu})$ based on Eq.~\ref{eq:forward}
\State Receive reward $r_t$ and new state $s_{t+1}$
\State $t \gets t + 1$
\Until terminal $s_t$ \textbf{or} $t-t_{start} ==$ rollout length $L$
\State Update $\phi$ in Eq.~\ref{eq:hw_diff} to acquire $\phi^*$
\For {$i \in \{t_{start},\ldots,t-1\}$}
\State $ \delta_{t} = r_{t} + \gamma V_{\pi}(s_{t+1} | \theta_{v}^{stu}) - V_{\pi}(s_t | \theta_{v}^{stu})$
\State Calculate $L_{task}$ in Eq.~\ref{eq:loss_total} based on $\delta_{t}$, $\pi(a_t,s_t | \theta_{\pi}^{tea})$, and $V_{\pi}(s_t | \theta_{v}^{tea})$
\State Calculate $L_{cost}$ in Eq.~\ref{eq:update_alpha} based on $\phi^*$
\State Update $\theta_{\pi}^{stu}$: $\theta_{\pi}^{stu} \gets \theta_{\pi}^{stu} - \eta_1 \nabla_{\theta_{\pi}^{stu}} L_{task}$
\State Update $\theta_{v}^{stu}$: $\theta_{v}^{stu} \gets \theta_{v}^{stu} - \eta_1 \nabla_{\theta_{v}^{stu}} L_{task}$
\State Update $\alpha$: $\alpha \gets \alpha - \eta_2 \nabla_{\alpha} (L_{task}+\lambda L_{cost})$
\EndFor
\Until $t > T_{max}$
\State Derive the final agent and accelerator with the highest $\alpha$ and $\phi$ respectively 

\noindent\Return the final agent and accelerator

\end{algorithmic}
\label{alg:a3c-s}
\end{algorithm}

\setlength{\textfloatsep}{4pt}

\begin{figure*}[b]
\begin{center}
\vspace{-1.5em}
   \includegraphics[width=0.95\linewidth]{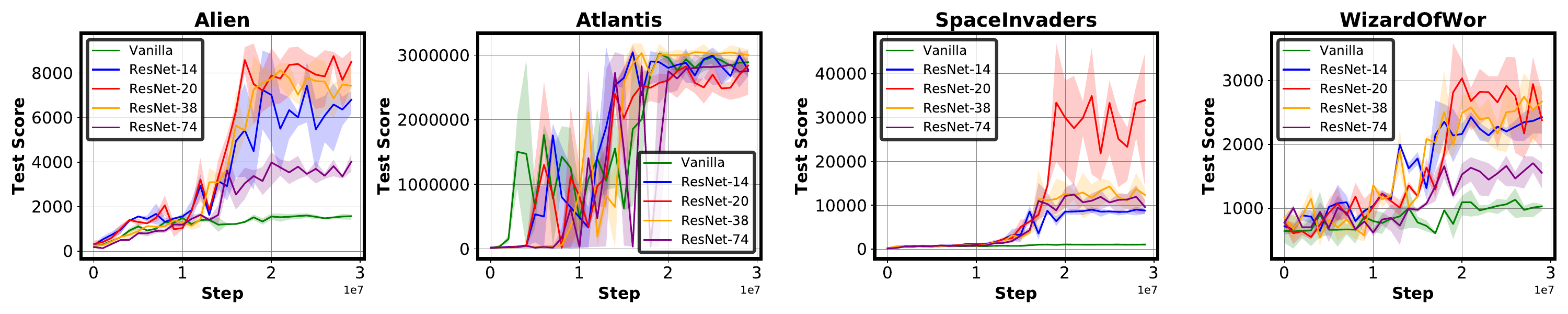}
\end{center}
  \vspace{-1.5em}
   \caption{Test scores averaged over 30 episodes during the training of five models on four Atari games.}.

\label{fig:scalability}
\end{figure*}

\section{Experiment results}

In this section, we first introduce our experiment setting, and then present ablation studies to evaluate A3C-S's component techniques and our A3C-S framework.

\subsection{Experiment setup}
\label{sec:exp_setup}

\textbf{Models and tasks.} We evaluate the performance of the AC-based DRL when its feature extractor backbone adopts the searched networks from A3C-S or five SOTA networks with different model sizes, including the original small network in DQN~\cite{mnih2015human} (termed as Vanilla) and ResNet-14/-20/-38/-74, on Atari 2600 games based on the Arcade Learning Environment. For all the ResNets, we modify the stride of the first convolution to be 2 and the output dimension of the final FC layer to be 256 to adapt them to the 84$\times$84 resolution of Atari games.  

\textbf{Training settings.} We use the same training and test hyper-parameters settings for all the models on all the tasks in this paper. Specifically, we train a DRL agent on each task for 3e7 steps with a discount factor ($\gamma$ in Eq.~\ref{eq:rl_policy_objective}) of 0.99 and a rollout length of 5; We use the RMSProp optimizer ~\cite{mnih2015human} with an initial learning of 1e-3 which keeps constant in the first 1e7 steps and then linearly decays to 1e-4; and the reported test score is averaged on 30 episodes with null-op starts following~\cite{mnih2015human}.

\textbf{Distillation Settings.} For the proposed AC-distillation, we train a ResNet-20 model as the teacher agent and $\beta_1$, $\beta_2$, and $\beta_3$ in Eq.~\ref{eq:loss_total} are set to be 1e-2, 1e-1, and 1e-3, respectively, in all the experiments.

\textbf{Search settings.} The supernet structure follows the network design (i.e., \#groups and stride) of the ResNet series with 12 sequential searchable cells. The candidate operators are standard convolutions with a kernel size 3/5, inverted residual blocks with a kernel size 3/5, a channel expansion of 1/3/5, and skip connections, leading to a search space of $9^{12}$ choices. We update the architecture parameters using an Adam optimizer with a momentum of 0.9 and a fixed learning rate of 1e-3. The initial temperature~\cite{wu2019fbnet} in Gumbel Softmax is set to 5 and decayed by 0.98 every 1e5 steps.

\begin{figure*}[!t]
\begin{center}
   \includegraphics[width=0.95\linewidth]{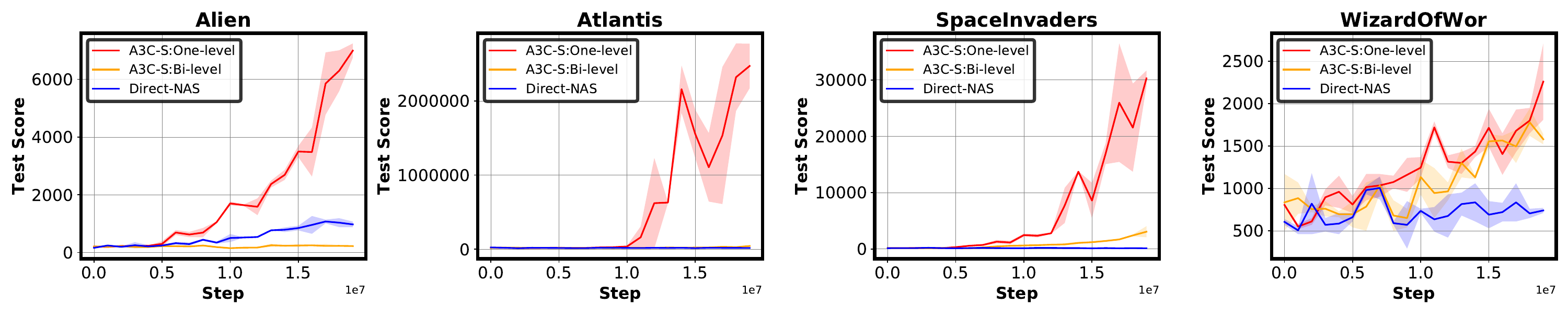}
\end{center}
  \vspace{-1.5em}
   \caption{Test score evolution during the search processes of three different search schemes on four Atari games~\cite{bellemare2013arcade}, where Direct-NAS denotes directly applying NAS w/o distillation, and A3C-S:One-level and A3C-S:Bi-level search with the distillation loss using one- and bi-level optimization, respectively.}
   \label{fig:nas}
   \vspace{-1.5em}
\end{figure*}

\textbf{Accelerator settings.} To evaluate A3C-S's generated accelerators, we adopt a standard FPGA design flow, i.e., the Vivado HLS flow~\cite{vivado_HLS}, and performance metric, i.e., frame per second (FPS). When searching for the accelerator parameter, A3C-S makes use of a SOTA accelerator performance predictor~\cite{xu2020autodnnchip,DNNCHIPPREDICTOR} to obtain fast and reliable estimation during search. The accelerator parameters we optimize upon include \textbf{1)} parallel processing elements (PE) settings: the number and inter-connections of PEs, 
\textbf{2)} buffer management: allocation of lower level memories between inputs, weights and outputs, \textbf{3)} tiling and scheduling of MAC (Multiply and Accumulate) computations, and \textbf{4)} layer allocation: ways to assign each layer to the corresponding pipeline stage (sub-accelerator).

\begin{table}[!t]
\vspace{-1.0em}
\caption{The test scores of different models on a subset of Atari games.}
  \resizebox{1.0\linewidth}{!}
  {    
\begin{tabular}{c|c|c|c|c|c}
\hline
Atari Games & Vanilla & ResNet-14 & ResNet-20 & ResNet-38 & ResNet-74 \\ \hline \hline
Breakout & 523.7 & 776.5 & 811 & \textbf{818.5} & 2.2 \\ 
Alien & 1724 & 9007 & \textbf{9323} & 8829 & 4456 \\ 
Asterix  & 4850 & 708500 & \textbf{856800} & 756120 & 539060 \\
Atlantis  & 3064320 & 3127390 & 3156130 & \textbf{3181090} & 3046490 \\ 
TimePilot  & 4780 & 9070 & \textbf{9680} & 9500 & 9040 \\ 
SpaceInvaders  & 1171 & 9848 & \textbf{46870} & 17962 & 15111 \\ 
WizardOfWor  & 1320 & 2690 & \textbf{3580} & 3160 & 1850 \\ 
Tennis  & -23.7 & 13.8 & 11.5 & \textbf{19.6} & 19.3 \\ 
Asteroids  & 2095 & \textbf{5690} & 5744 & 1947 & 4792 \\ 
Assault  & 10164 & 14470 & \textbf{17314} & 12406.5 & 9849 \\ 
BattleZone  & 7600 & 5800 & 13100 & \textbf{13300} & 4100 \\ 
BeamRider  & 5530 & 23984 & 25961 & 29498 & \textbf{30048} \\ 
Bowling  & 28.1 & 53 & \textbf{59.2} & 33.2 & 50.8 \\
Boxing  & 4.2 & \textbf{100} & \textbf{100} & 99.3 & 87.1 \\
Centipede  & 5025 & 6690 & 6410 & 6384.6 & \textbf{6899} \\ 
ChopperCommand  & 1320 & 11170 & \textbf{14910} & 4370 & 8240 \\ \hline
\end{tabular}
    }
  \label{tab:scalability}
\end{table}

\subsection{Ablation study: DRL with different model sizes}
\label{sec:exp_scalability}

\textbf{Observations and Analysis.} Fig.~\ref{fig:scalability} visualizes the test core evolution during the training process on various Atari games, when adopting different networks for the DRL agents, where the highest achieved test scores is listed in Tab.~\ref{tab:scalability}. Two observations can be made. \underline{First}, networks with a larger size in general favor the achieved test scores especially on more difficult tasks (e.g., BeamRider of Tab.~\ref{tab:scalability}), since larger networks can provide higher test scores with the same training time steps in most of the games over the smaller vanilla-network and ResNet-14; \underline{Second},
there always exists a task-specific optimal network size (i.e., a further increase won't improve or even degrade the test score), which is likely due to the increased difficulty of training larger DRL agents. For instance, the vanilla network performs well in the Atlantis game (see Fig.~\ref{fig:scalability}), whereas ResNet-38 merely offers a marginally improved score at a cost of 13.7$\times$ higher FLOPs (floating-point operations); and ResNet-74 is inferior to ResNet-20/38 in most of these experiments, since it is more difficult to be trained within the limited training steps. Note that even though more training steps or better-tuned hyper-parameters may improve the convergence of DRL with ResNet-74, the associated inefficiency makes it impractical to be widely adopted.

\begin{table*}[b]
\vspace{-1em}
 \centering
\caption{DRL with the vanilla network and ResNet-14 under: (1) no distillation, (2) only policy distillation, and (3) our AC-distillation.}
  \resizebox{0.85\linewidth}{!}
  {    
\begin{tabular}{c|ccc|ccc}
\hline
& \multicolumn{3}{c}{ Vanilla } & \multicolumn{3}{c}{ ResNet-14 } \\ \hline \hline
Atari Games & No distillation & Policy disitllation only & AC-distillation & No distillation & Policy disitllation only & AC-distillation \\  \hline
Alien  & 1724 & 3096 & \textbf{3419} & 9007 & 14682 & \textbf{15723} \\
SpaceInvaders  & 1171 & 26821 & \textbf{30124} & 9848 & 76246 & \textbf{111189} \\ 
Asterix  & 4850 & 59020 & \textbf{64510} & 708500 & 749870 & \textbf{849400} \\ 
Asteroids  & 2095 & 4131 & \textbf{4647} & 5690 & 15371 & \textbf{15947} \\ 
Assault  & 10164 & 8088.4 & \textbf{9628.5} & 14470 & 11697 & \textbf{14052} \\ 
BattleZone  & 7600 & 14200 & \textbf{14400} & 5800 & 16300 & \textbf{17500} \\ 
BeamRider  & 5530 & 14417 & \textbf{21519} & 23984 & 38311 & \textbf{39604} \\ 
Boxing  & 4.2 & 2.8 & \textbf{100} & \textbf{100} &\textbf{100} & \textbf{100} \\
Centipede  & 5025 & 5800 & \textbf{6575.5} & 6690 & 7744.3 & \textbf{8056.9} \\ 
ChopperCommand  & 1320 & 15900 & \textbf{19120} & 11170 & 26320 & \textbf{31190} \\ 
CrazyClimber  & 118300 & 138610 & \textbf{145700} & 128710 & 135290 & \textbf{138470} \\ 
DemonAttack  & 318349 & 463823 & \textbf{483490} & 481818 & 517801 & \textbf{521051} \\ \hline
\end{tabular}
    }
  \label{tab:distillation}
\end{table*}

\textbf{Extracted Insights.} The observations above imply that (1) DNNs' architecture plays a critical role in DRL, which is still under-explored in existing works, and (2) designing task-specific agents is highly desired in optimally balancing the test score and processing efficiency for different tasks. Given that the manual design of dedicated agents for different tasks is not practical in handling the growing need for the fast development of DRL-powered solutions for numerous applications, we are motivated to use NAS to design DRL agents.

\subsection{Ablation study: evaluating the proposed AC-distillation}
\label{sec:exp_distillation}

\textbf{Observations and Analysis.}
We compare our AC-distillation mechanism with the three baselines by applying them to the vanilla network and ResNet-14 evaluated on Atari games (see Tab.~\ref{tab:distillation}). Three observations can be made. \underline{First}, compared to designs without distillation, distillation strategies in general favor the test scores, which is consistent with~\cite{rusu2015policy}; 
\underline{Second}, our AC-distillation mechanism consistently performs best in achieving the highest test scores on most tasks, among the three distillation strategies.


\subsection{Ablation study: one-level vs. bi-level optimization}
\label{sec:exp_nas}

In this set of experiments, we visualize the test score evolution in Fig.~\ref{fig:nas} during the search process of (1) directly applying NAS without distillation, (2) using AC-distillation with bi-level optimization, and (3) using AC-distillation with one-level optimization. 
We can see that the test scores remain low when searching with the bi-level optimization, validating our mentioned hypothesis that the supernet cannot serve as an accurate proxy to indicate the performance of the sampled subnetworks. In contrast, searching with the one-level optimization leads to a consistent improvement in the test scores during search, demonstrating the first framework that successfully makes NAS possible in DRL.



\begin{figure*}[!tbh]
\begin{center}
   \includegraphics[width=0.95\linewidth]{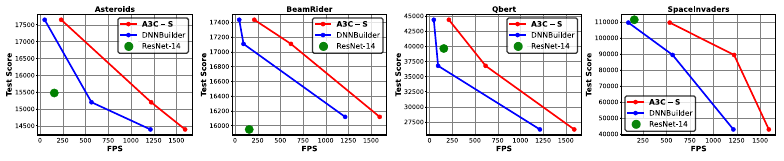}
\end{center}
  \vspace{-2em}
   \caption{Benchmark the proposed A3C-S with (1) ResNet-14 on our DAS's searched accelerators and (2) A3C-S searched agents on A3C-S searched accelerators vs. SOTA accelerators DNNBuilder~\cite{zhang2018dnnbuilder}, in terms of test scores and FPS trade-off on four Atari games.}
   \label{fig:sota}
   \vspace{-1em}
\end{figure*}

\subsection{Evaluating the proposed A3C-S framework}

\textbf{A3C-S vs. SOTA networks on searched accelerators.} Here we evaluate A3C-S's searched agents by comparing their hardware performance (i.e., FPS) with the most competitive SOTA DRL agents, i.e., ResNet-14 with the best trade-off between test scores and hardware efficiency among all our experiments (see Tab.~\ref{tab:distillation}). For a fair comparison, we use our proposed DAS engine to search for the optimal accelerators under the same search settings and to train both networks with our AC-distillation mechanism. From Fig.~\ref{fig:sota}, we can see that the resulting hardware efficiency of A3C-S's searched agents is consistently higher than that of the most competitive SOTA DRL agents, under a comparable or better test score. This set of experiments again motivates the necessity of applying NAS to search for task-specific optimal agents that balance both the test scores and model complexity.

\textbf{A3C-S searched agents on A3C-S searched accelerators vs. SOTA accelerators.}
Here we evaluate A3C-S's DAS engine by comparing the hardware efficiency (i.e., FPS) of A3C-S's searched agents, when being accelerated by both our DAS's generated accelerators and a SOTA DNN accelerator, DNNBuilder \cite{zhang2018dnnbuilder}. For a fair comparison, both accelerators adopt the optimal DRL agents searched by our A3C-S. As shown in Fig.~\ref{fig:sota}, the FPS achieved by DAS's generated accelerators consistently outperforms that of the SOTA DNN accelerator, under the same DSP limit (900; which is the largest resource in our ZC706~\cite{zc706}. This set of experiments validates the need for dedicated accelerators for DRL agents, of which our A3C-S is the first demonstration.

\textbf{A3C-S vs. SOTA DRL solutions.} To evaluate the combined benefits of our A3C-S, we evaluate A3C-S's resulting accelerators (i.e., using A3C-S's DAS to generate optimal accelerators for A3C-S's searched agents) over a SOTA DRL system FA3C~\cite{cho2019fa3c}, where the performance of the latter is directly obtained from the reported data of the baseline paper. From Tab.~\ref{tab:benchmark_fa3c}, we can see that our A3C-S's resulting DRL accelerators achieve 2.1$\times$ $\sim$ 6.1$\times$ better FPS, while offering consistently higher test scores, as compared to the SOTA DRL system~\cite{cho2019fa3c}. As expected, jointly searching for both the DRL agents and their accelerators leads to optimal DRL solutions, achieving both higher test scores and better hardware efficiency. Furthermore, our A3C-S's differentiable search strategy makes it more accessible to researchers without paramount computing resources and facilitates fast development of DRL-powered solutions.

\begin{table}[!h]
\vspace{-1.5em}
\centering
\caption{\textbf{Test scores / FPS} of our A3C-S compared with FA3C on six Atari games reported by FA3C~\cite{cho2019fa3c}.}
\vspace{-1.em}
  \resizebox{0.65\linewidth}{!}
  {    
\begin{tabular}{c|c|c}
\hline
Atari Games & FA3C & \textbf{A3C-S}\\ \hline
BeamRider & 3100 / 260  & 36745 / 617.7  \\
Breakout & 340 / 260 & 670 / 1596.3 \\ 
Pong & 0 / 260 & 20.9 / 787.4 \\ 
Qbert  & 6100 / 260 & 15194 / 1222.9 \\
Seaquest  & 170 / 260 & 478940 / 778.1 \\
SpaceInvaders & 830 / 260  & 109417 / 535.6 \\ \hline
\end{tabular}
    }
  \label{tab:benchmark_fa3c}
\end{table}


\vspace{-0.2cm}
\section{Conclusion}
\vspace{-0.1cm}

We propose, design, and validate a DRL agent accelerator co-search framework dubbed A3C-S, which to our best knowledge is the first to (1) automatically co-search for the optimally matched DRL agents and accelerators that maximize both test scores and hardware efficiency, and (2) demonstrate the first successful DNAS for DRL, for which a vanilla DNAS fails due to DRL's training instability. 

\section*{Acknowledgements}
The work is supported by the National Science Foundation (NSF) CCRI-2016727 and CAREER-2048183 Awards.

\bibliography{ref}
\bibliographystyle{IEEEtran}
\end{document}